# Contribución de la semántica combinatoria al desarrollo de herramientas digitales multilingües

María José Domínguez Vázquez [1]

**Resumen.** El presente trabajo describe cómo la semántica combinatoria ha contribuido al diseño de tres prototipos de generación automática de esquemas argumentales del nombre en español, francés y alemán – *Xera*, Combinatoria y CombiContext. Asimismo, da cuenta de la importancia del conocimiento sobre la interfaz sintáctico-semántica argumental en una situación de producción en lenguas extranjeras. Tras un apartado descriptivo sobre el diseño, tipología y niveles informativos de los recursos, sigue una exposición sobre el papel fundamental del significado combinatorio –relacional y categorial– en su desarrollo. El estudio detalla diferentes filtros semánticos aplicados en la selección, organización y expansión del caudal léxico, siendo estos piezas clave para la generación de frases nominales mono y biargumentales gramaticalmente correctas y semánticamente aceptables.
**Palabras clave:** significado combinatorio; Generación automática de la lengua; roles semánticos; rasgos ontológicos; esquemas argumentales

## [en] Contribution of the Combinatorial Semantics to the development of multilingual digital tools

**Abstract.** This paper describes how the field of Combinatorial Semantics has contributed to the design of three prototypes for the automatic generation of argument patterns in nominal phrases in Spanish, French and German – *Xera*, Combinatoria and CombiContext. It also shows the importance of knowing about the argument syntactic-semantic interface in a production situation in the context of foreign languages. After a descriptive section on the design, typologie and information levels of the resources, there follows an explanation of the central role of the combinatorial meaning – roles and ontological features. The study deals with different semantic filters applied in the selection, organization and expansion of the lexicon, being these key pieces for the generation of grammatically correct and semantically acceptable mono- and biargumental nominal phrases.
**Keywords:** combinatory meaning; Automatic Language Generation; semantic roles; ontological features; argument patterns

**Índice.** 1. Introducción. 2. Datos generales e interfaz de consulta de los generadores. 2.1. Punto de partida. 2.2. Descripción general y metodología. 2.3. Aplicación web de consulta. 3. Los generadores automáticos en el panorama lexicográfico: tipología. 4. De la interfaz sintáctico-semántica a la generación automática. 4.1. El esquema argumental como punto de partida. 4.2. El significado combinatorio. 4.3. Los roles semánticos. 4.3.1. Los roles semánticos como desambiguadores de significado. 4.3.2. Los roles semánticos como instancias grupales. 4.3.3. A modo de resumen. 4.4. Rasgos ontológicos como componentes del significado combinatorio: prototipos léxicos y clases semánticas. 5. Conclusión y perspectivas de futuro. Agradecimientos. Referencias.

# 1. Introducción

En la era de la inteligencia artificial y de la comunicación digital, el desarrollo de innovadores recursos lingüísticos y lexicográficos de calidad implica potenciar la integración y retroalimentación de datos, así como la automatización de técnicas y procedimientos de extracción, procesamiento y generación del lenguaje natural. En esta línea, un análisis de las publicaciones recogidas en *OBELEXmeta* –recurso que compila literatura científica en el campo de la lexicografía electrónica– permite constatar que desde los años 80 se establece una mayor interacción entre las áreas de la lexicografía y del procesamiento del lenguaje natural (PLN) y generación (GLN). De hecho, hace unas décadas no contaríamos con estudios como *Lexicography and Natural Language Processing* (Horák y Rambousek, 2018), *Lexicography between NLP and Linguistics: Aspects of Theory and Practice* (Trap-Jensen, 2018) o *Natural Language Processing and Automatic Knowledge Extraction for Lexicography* (Krek, 2019).

1 Universidade de Santiago de Compostela–Instituto da Lingua Galega. Correo electrónico: majo.dominguez@usc.es ORCID: https://orcid.org/0000-0002-6060-9577

Es innegable que la Lexicografía y la Lingüística se benefician de los avances en la generación y el procesamiento del lenguaje natural - extracción de datos automatizada (Kilgarriff y Kosem, 2012), medidas de similitud semántica mediante métodos predictivos como *word2vec* (Mikolov et al, 2013) o *fastText* (Bojanowski et al., 2017) o generación automática de diccionarios (Bardanca, 2020). Y a la inversa: las áreas del PLN han demostrado beneficiarse de la disponibilidad de conocimiento léxico a diferentes niveles (Navigli y Ponzeto, 2012; Trap-Jensen, 2018). Además contamos, entre otros, con estudios más pormenorizados sobre el uso y los usuarios del diccionario y los recursos digitales (Müller-Spitzer, 2014; Müller-Spitzer et al, 2018), sobre el usuario como consumidor-"prosumidor" de información –en especial en la lexicografía colaborativa (Meyer y Gurevych, 2012)–, sobre el manejo de nuevos dispositivos y accesos a la información (Prinsloo et al, 2011, 2012; Nied, 2014) y sobre diferentes tipos de consulta, necesidades y expectativas de los usuarios (Gouws, 2017). Estas interacciones, así como los avances en las tecnologías de la lengua y el empoderamiento del usuario de los recursos lexicográficos han influido notablemente en el propio concepto de diccionario y en el tipo de tareas que la lexicografía ha de acometer (Maldonado, 2019; tesis de Villa Vigoni, 2018)

Precisamente en esta nueva concepción del diccionario del futuro – entendido como sistema de información en la línea de las citadas tesis de Villa Vigoni – se enmarcan nuestros prototipos de generación automática argumental del nombre en español, francés y alemán, *Xera, Combinatoria y CombiContext* (http://portlex.usc.gal/combinatoria/). Estos sirven de modelo para un nuevo tipo de diccionario plurilingüe de valencias, que quiere dar respuesta a las demandas de los usuarios - consultas sencillas y rápidas, así como la obtención individualizada y concreta de datos (Spohr, 2011). Seguimos pues la máxima de que los recursos se desarrollan para ser consultados, en consonancia con la relevancia otorgada al usuario de la información lexicográfica en la literatura científica.

El artículo se enmarca, por tanto, en las aproximaciones previamente señaladas, y tiene como objetivo presentar el papel que juega el significado combinatorio – relacional y categorial (Engel, 2004) – como elemento nuclear del nuevo método combinado que se ha aplicado en el desarrollo de los generadores (apartado 4). Dicho método permite describir, procesar y generar sintáctica y semánticamente el potencial combinatorio de la frase nominal en el eje sintagmático y paradigmático frasal (la valencia activa) y en el oracional (la valencia pasiva). Para tal fin, se analizan los patrones argumentales y el significado combinatorio – roles semánticos y rasgos ontológicos (4.2. y 4.3.) – de los elementos implicados en una expresión, así como los prototipos léxicos y clases semánticas (véase 4.4.) actualizables en las diferentes casillas funcionales. De este modo, el modelo y, en definitiva, los generadores cuentan con un sólido fundamento semántico. Por tanto, nuestros prototipos se asientan en diferentes aproximaciones lingüísticas, así como en procedimientos de recuperación y extracción de datos mediante técnicas de PLN y, finalmente, en la generación automática (GLN).

Esta contribución se articula como sigue: En los apartados iniciales se aporta una visión de conjunto de los simuladores *Combinatoria* y *CombiContext* (apartado 2) junto con su descripción tipológica (apartado 3). El apartado 4 se centra en la aplicación de la semántica combinatoria en el desarrollo de los generadores.

## 2. Datos generales e interfaz de consulta de los generadores

### 2.1. Punto de partida

Hoy en día están en funcionamiento tres generadores automáticos de patrones argumentales y sus ejemplos: *Xera* aporta esquemas monoargumentales de la frase nominal y *Combinatoria* esquemas biargumentales. *CombiContext* permite recrear el contexto frasal y oracional de las frases nominales generadas por los simuladores previamente citados. Dichos recursos surgen ligados a una necesidad doble y, por tanto, cuentan también con una doble aplicación:

- El origen de nuestros generadores son las constatadas dificultades a la hora de compilar ejemplos de corpus adecuados para mostrar la combinatoria valencial de las cinco lenguas recogidas en el diccionario multilingüe de la valencia del nombre *Portlex* (Domínguez y Valcárcel, 2020). Dado que los corpus manejados no están anotados semánticamente, no es posible extraer ejemplos que atiendan a la interfaz sintáctico-semántica valencial, esto es, que reflejen el esquema sintáctico-semántico valencial y el potencial combinatorio de las unidades de análisis (*vid.* 3.). Por tanto, los búsquedas de corpus no permiten diferenciar realizaciones argumentales-valenciales frente a aquellas que no lo son (*vid.* 4.3.1.). Por este motivo, decidimos invertir la perspectiva: generar automáticamente esquemas argumentales y ejemplos adecuados para nuestro diccionario, en lugar de extraerlos. A diferencia de los corpus, los generadores también permiten extraer y consultar los datos atendiendo a criterios semántico-ontológicos.
- Diferentes estudios señalan que la adquisición de un lexema está ligado al aprendizaje de su entorno sintáctico-semántico (Laufer y Nation, 2012) y que un nutrido número de errores en la producción en lenguas extranjeras se sitúa precisamente aquí (Gao y Haitao, 2020; Nied, 2014). Dado que, además, se constata una carencia de recursos electrónicos plurilingües sobre la valencia nominal, estos simuladores vienen a cubrir una laguna en la enseñanza-aprendizaje de lenguas, en donde se recurre a diferentes aplicaciones didácticas de los recursos lexicográficos (Alvarez de la Granja, 2017; Müller-Spitzer et al, 2018; Wolfer et al., 2016).

## 2.2. Descripción general y metodología

Los generadores piloto ofrecen información de 20 sustantivos de diferentes campos semánticos en alemán, español y francés, los cuales recoge la Figura 1:

Figura 1. Sustantivos y campos semánticos

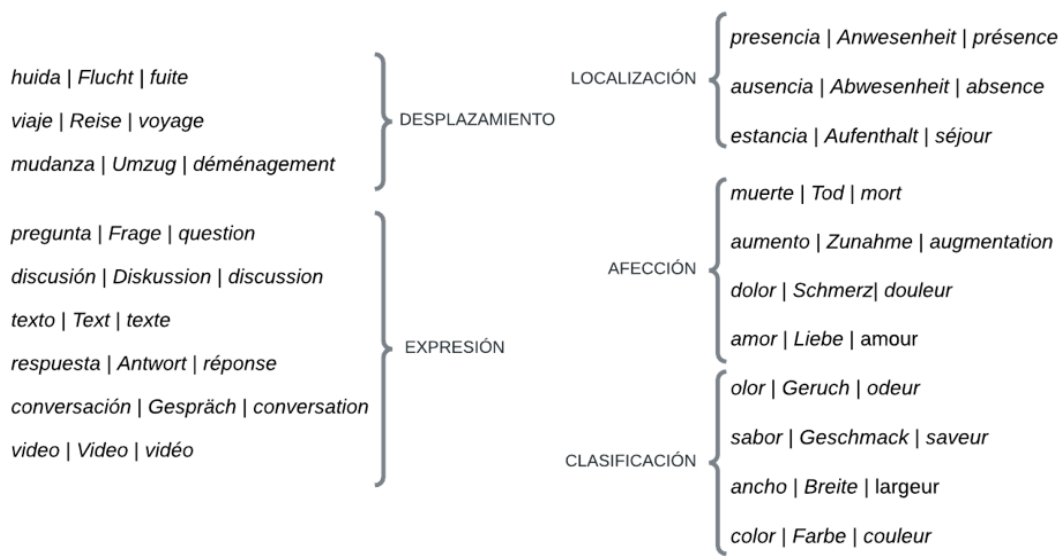

El primer filtro para la inclusión de los sustantivos en los generadores es su estatus como portadores valenciales, esto es, su capacidad de abrir casillas funcionales (Engel, 2004). Nuestro modelo incluye la descripción de sustantivos derivados, por ejemplo, *huida*, sustantivos no deverbales, como *presencia*, e incluso aquellos que no muestran ningún tipo de derivación, como *texto* o *vídeo*. Contemplamos, a su vez, un segundo criterio: la pertenencia de los sustantivos a diferentes escenarios (Fillmore, 1977), la cual determinamos a partir del análisis de los diferentes roles semánticos o relatores implicados en el esquema argumental (*vid*. 4.2.).

Tras la selección de los sustantivos objeto de análisis, describimos su esquema o patrón argumental (*vid*. 4.1.) siguiendo los presupuestos teórico-valenciales del diccionario Portlex (2018). A continuación, es necesario determinar qué caudal léxico puede cubrir el eje paradigmático de los diferentes complementos específicos, para, de este modo, poder codificar la generación de las realizaciones argumentales. Así, en relación con el patrón argumental del sustantivo esp. *sabor* [*el sabor de algo a algo*] se necesita responder a la pregunta de qué unidades léxicas pueden ocupar ambas casillas, como, por ejemplo, *el sabor del* {*asado* | *pescado* | *churrasco, ...*} *a* {*ajo* | *cebolla* | *pimienta* | *alcohol,* ...}. Para tal fin, extraemos datos de frecuencia y combinatoria de *Sketch Engine*. Los clasificamos según su significado relacional y realizamos un prototipado semántico en dos niveles: establecemos, en primer lugar, los prototipos o candidatos léxicos (Domínguez, 2021) y, en segundo lugar, las clases semánticas específicas (*vid*. 4.4.).

Consideramos prototipos o candidatos léxicos aquellos lexemas o unidades léxicas que ocupan típicamente una determinada casilla de una estructura valencial concreta de un núcleo. No se trata de conceptos abstractos o categorizaciones cognitivas (cfr. Rosch, 1978), dado que nuestra noción de prototipo está ligada a la estructura argumental sintáctica de un núcleo nominal concreto y la actualización de su(s) posible(s) lectura(s) o acepciones. Su tipicidad, así como su representatividad, se basa en la interacción de dos parámetros: la realización de un rol semántico *vid*. (4.3.) y su frecuencia en dicha estructura sintáctico-semántico argumental. Por tanto, el objetivo no es determinar si {cabeza} o {pelo} son *per se* prototipos centrales de la clase semántica {parte del cuerpo humano}, sino si estos lo son en una determinada estructura argumental. Así, para el patrón sintáctico [*EL* + *DOLOR* + DE + {N}] podemos concluir que {cabeza} – ej. *el dolor de cabeza* – es un prototipo léxico, frente a {pelo} que no es representativo de dicho patrón. Por el contrario, en el esquema argumental [*EL* + *COLOR* + *DE* + {N}], {pelo} sí que se puede considerar un candidato léxico prototípico – ej. *el color de pelo*. Ya en este nivel descriptivo el significado combinatorio, tanto en el plano relacional como categorial, es central *vid*. (4.4.).

Para el análisis ontológico de los candidatos léxicos aplicamos una ontología léxica (Domínguez, Valcárcel y Bardanca, 2021), la cual compendia clases semánticas prototípicas resultantes de la vinculación entre los diferentes sustantivos y su(s) estructura(s) argumental(es) concreta(s). De este modo, se puede establecer el paradigma de clases semánticas y unidades léxicas de una determinada casilla funcional, lo cual resulta relevante desde un punto de vista lingüístico e imprescindible en el procesamiento de extracción y generación automática de datos. Así, una vez delimitadas las unidades léxicas y clases semánticas que acostumbran a ocupar una casilla funcional concreta, tiene lugar el proceso de expansión léxica en el eje paradigmático: mediante los rasgos ontológicos adjudicados a los prototipos léxicos y las clases semánticas accedemos al caudal léxico recogido en *WordNet* (Gómez y Solla, 2018), organizado en diferentes ontologías. De este modo, extraemos un conjunto más amplio de candidatos léxicos representativos, los cuales comparten las características semánticas de los prototipos léxicos de los que partimos. Los datos extraídos siguen diferentes fases de depurado, tratamiento y empaquetado paradigmático, finalizándose con su generación. Para tal fin, se maneja en el caso de las estructuras biargumentales un conjunto de herramientas de creación propia (para una descripción detallada del procedimiento, así como de las herramientas manejadas véase Domínguez, Solla y Valcárcel, 2019). La generación de las estructuras oracionales se asienta en un *PoS(Part of Speech) tagger* de diseño propio, que extrae datos de la Wikipedia.

Las diferencias entre los generadores son de diferente calado: *Xera* ofrece estructuras monoargumentales de la frase nominal con un acceso inicial a los datos de tipo formal. A diferencia de este generador, el segundo simulador, *Combinatoria*, presenta la estructura biargumental de la frase nominal con una estructura de acceso conceptual (*vid.* 2.3.). El último de los generadores desarrollados, *CombiContext*, bebe de las fuentes de los anteriores y recrea el marco oracional y frasal en el que se pueden realizar las frases nominales generadas por los simuladores previos.

Actualmente la base de datos que sustenta a los *madre/madres* generadores cuenta con 90 500 formas y 45 000 lemas, asociados a diferentes clases semánticas. Por ejemplo, *madre/madres* son formas de representantes léxicos de la categoría semántica {animado humano familia} y *madre* es un lema de dicha categoría. Además, se han analizado 3600 argumentos, 20 600 estructuras sintáctico-semánticas y más de 85 000 estructuras oracionales (*vid.* Figura 2).

Figura 2. Representación cuantitativa y cualitativa de niveles descriptivos

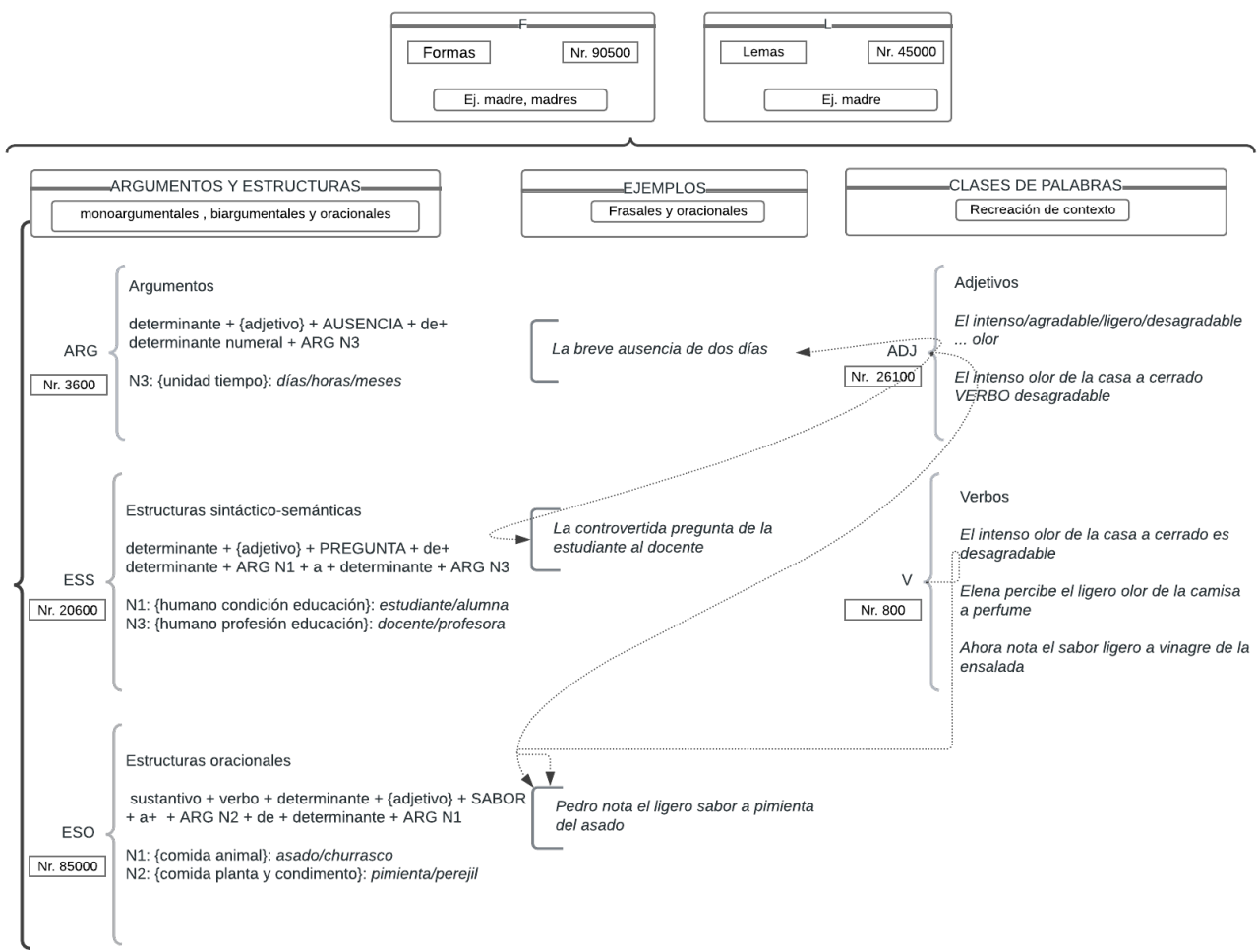

Cabe señalar que en *CombiContext* un sustantivo cuenta de media con 83 estructuras oracionales (ESV) de tipo formal, por ejemplo: (a) sujeto[Frase nominal] + verbo + adverbio: *El viaje de Mario a Berlín termina así*; (b) sujeto[Frase nominal] + verbo + atributo: *La fuga de los refugiados desde la frontera es peligrosa*; (c) adverbio + sujeto + verbo + objeto directo sujeto[Frase nominal]: *Ahora Carlos nota el olor a humedad de la casa;* (d) sujeto + verbo + adverbio + suplemento[Frase nominal: *Nerea escribe brevemente sobre el amor de Marco Antonio por Cleopatra*, etc. Esta cantidad de ESVs se refiere únicamente a estructuras formales, sin tener en cuenta toda la variabilidad paradigmática de las clases semánticas combinables entre sí y los diferentes verbos, adverbios o adjetivos que pueden ser actualizados. Lo mismo sucede con las estructuras sintáctico-semánticas (ESS), 343 estructuras de media por sustantivo.

## 2.3. Aplicación web de consulta

El principal cometido de las herramientas consiste en dar cuenta de las interacciones argumentales sintáctico-semánticas del nombre. Así, por ejemplo, en la interfaz de la herramienta *Combinatoria* (Figura 3), un usuario selecciona de un desplegable i) la lengua, ii) el sustantivo, iii) la combinación binaria y distribucional de los argumentos junto con su(s) clase(s) semánticas (*vid.* 4.4.) y iv) la estructura. La aproximación conceptual de esta herramienta – a diferencia de *Xera* – y la presencia de ventanas emergentes están en consonancia con estrategias de didactización de los recursos (*vid.* 3.), las cuales resultan de experimentos previos realizados con usuarios de diferente tipo.

Figura 3. Herramienta *Combinatoria*

Dado que el recurso presenta argumentos específicos del nombre, el orden de aparición de las unidades en la secuencia es relevante. Así, por ejemplo, en el ejemplo (1) *de Pedro* es interpretable como el dueño o habitante habitual de la casa y, por tanto, podríamos contar con una lectura como (2). El ejemplo (3) no deja lugar a posibles interpretaciones: *Pedro* es el que realiza la estancia, y, por tanto, desempeña –a diferencia de (1)– un rol semántico.

(1)  *La estancia en la casa de Pedro*
(2)  *La estancia de Juan en la casa de Pedro*
(3)  *La estancia de Pedro en la casa*

Realizada la elección, el usuario cliquea en la pestaña *generar frases* y obtiene resultados, también exportables en los formatos JSON y CSV (*vid*. Figura 4):

Figura 4. Ejemplos generados automáticamente en español con la herramienta *Combinatoria* para [olor a {líquido no consumible} + de + determinante + {lugar construcción habitación}]

| frases generadas |
| --- |
| el olor a aguarrás de las solanas |
| el olor a espral del campanario |
| el olor a cicuta de los anfiteatros |
| el olor a resina de los desvanes |
| el olor a lejía de la habitación |
| el olor a suavizante de los aseos |
| el olor a aguarrás de los urinarios |

| el olor a pesticida de las buhardillas |
| --- |
| el olor a agua oxigenada de la sala de baile |
| el olor a alcohol de los zaguanes |
| el olor a aguarrás de los campanarios |
| el olor a agua salada de las salas de billar |
| el olor a disolvente del trastero |
| el olor a agua salada de los compartimentos |
| el olor a resina de los vestíbulos |

La generación automática de los datos resulta de una aleatoria restringida, lo que supone que esta afecta a los representantes léxicos, no al rol ni a la clase semántica (*vid*. 4.). Esto es así, en primer lugar, porque el propio usuario selecciona previamente la estructura argumental y la clase semántica; en segundo lugar, porque las clases semánticas están preestablecidas para cada sustantivo (*vid*. 4.4.). Además, la herramienta ofrece un filtrado de la combinatoria sintagmática argumental mediante el empleo de *word embeddings* –representaciones vectoriales de una palabra en contexto. Para tal fin, aplicamos los métodos predictivos *word2vec (*Mikolov et al, 2013) *y fastText* (Bojanowski et al., 2017). Una de sus implementaciones –CBOW– intenta predecir qué palabra resulta más adecuada para un contexto específico, esto es, qué palabra ocupa frecuentemente una casilla (Bardanca Outeiriño, 2020). Obtenemos, por tanto, datos filtrados según su frecuencia de coaparición en contexto (*vid*. Figura 5), tratándose, en definitiva, de un filtrado semántico-distribucional en el sentido de Engel (2004: 188) o Firth (1957: 11). La aplicación de *fastText* (Figura 5) conlleva un volumen menor de datos y una menor variabilidad en cuanto a los candidatos léxicos combinados, a diferencia de los datos obtenidos sin su aplicación (Figura 4). En este último caso, los resultados son más números y cuentan con mayor variabilidad. Si bien cumplen criterios de gramaticalidad, algunas combinatorias pueden resultar poco aceptables o coherentes atendiendo a nuestro conocimiento enciclopédico o cultural.

Figura 5. Datos de combinatoria sintagmática aplicando *fastText* en fr. [determinante + MORT + DE + determinante + actante N1{animado humano familia} + PAR + actante N2 {proceso natural patológico}]

☑ Filtrar con Word2Vec

límite de frases :880

| GENERAR FRASES | EXPORTAR FRASES EN JSON | EXPORTAR FRASES EN CSV |
| --- | --- | --- |

| frases generadas |
| --- |
| la mort du nourrisson par infection |
| la mort du nouveau-né par complications infectieuses |
| la mort du nourrisson par complications chirurgicales |
| la mort du nouveau-né par éclampsie |
| la mort du nouveau-né par complications cardiaques |
| la mort de la nouveau-née par pneumonie |
| la mort du nourrisson par sepsis |
| la mort de la nouveau-née par complications chirurgicales |
| la mort du nourrisson par tuberculose |
| la mort du nourrisson par botulisme |
| la mort du nouveau-né par complications chirurgicales |
| la mort de la nouveau-née par complications infectieuses |

*CombiContext*, la herramienta para la recreación contextual y frasal de las frases nominales generadas, presenta también una aproximación distribucional en el primer acceso a su interfaz de consulta (Figura 6): es posible establecer si el verbo aparece antes o después de la frase nominal. El siguiente filtro permite seleccionar esquemas formales básicos, a los que acompañan ejemplos estándar para mayor claridad. Clicando en el esquema formal que se desea consultar se obtienen estructuras oracionales con información semántica sobre sus diferentes casillas valenciales.

Figura 6. Interfaz 1.0. de *CombiContext*

## 3. Los generadores automáticos en el panorama lexicográfico: tipología

Los simuladores proponen un nuevo concepto de diccionario plurilingüe – en este caso de diccionario valencial automático e interactivo (Prinsloo et al, 2011) integrado en el portal lexicográfico PORTLEX. Atendiendo a sus características mediales, estos recursos multilingües se conciben desde un principio como recursos en línea accesibles de modo simultáneo por más de un usuario. Por su carácter de prototipos o simuladores se encuentran en constante actualización, además son de acceso libre y gratuito. Han sido diseñados como herramientas lexicográficas independientes de consulta para un destinatario humano, pero integrables en otro tipo de recursos y exportables como léxicos computacionales (compárese Kabashi, 2018: 855). Los generadores conceden a las tres lenguas descritas el mismo estatus descriptivo en cuanto a la cantidad y tipo de información aportada.

Dichos prototipos muestran similitudes y divergencias tipológicas con recursos de su entorno más cercano:

a) Contamos con herramientas que generan automáticamente diferentes tipos de textos (Nallapati, Zhou, Nogueira dos Santos, Gulcehre y Xiang, 2016; Sordoni et al., 2015), frases literarias (Moreno Jiménez, Torres-Moreno, Wedemann y SanJuan, 2020), textos a partir de imágenes y viceversa (Otter, Medina y Kalita, 2019) y, en algunos casos, chistes, poemas o historias casi sin *input* de partida (Roemmele, 2016). Estos persiguen, sin embargo, finalidades distintas a nuestros simuladores.

b) Difieren también de otros recursos que sí ofrecen datos sobre la interfaz sintáctico-semántica, si bien, en especial, para el inglés y para el verbo como unidad de análisis. Sirven aquí como ejemplo *Verbnet*, *Propbank*, *CPA* o *Framenet*; ejemplos para el inglés en español son *Verbario* y *AnCora*. No obstante, existen taxonomías para el sustantivo, como *Kind*, y proyectos como *AncoraNom* o *NomBank*. Estos últimos no describen, a diferencia de nuestros simuladores, las diferentes clases semánticas asociadas a un rol determinado ni los prototipos léxicos (*vid*. 4), tampoco lo contemplan entre sus objetivos.

c) Diferentes estudios versan sobre la generación automática de contenidos lexicográficos: diccionarios (Bardanca, 2020; Kabashi, 2018; Delli, Bovi y Navigli, 2017), artículos lexicográficos (Geyken, Wiegand y Würzner, 2017) o partes del diccionario (los ejemplos, en Kosem et al, 2019). Estos no pretenden, sin embargo, generar automáticamente ejemplos de esquemas argumentales nominales según criterios de selección establecidos por el usuario.

d) Su clasificación como nuevo tipo de diccionario valencial remite a la teoría gramatical en el que se sustentan y al objeto que describen: la estructura y combinatoria argumental de un portador valencial. Junto con su carácter plurilingüe y electrónico, existen además significativas diferencias con otros diccionarios de valencias: nuestros simuladores se perfilan como un modelo para nuevos diccionarios valenciales automáticos e interactivos, que pueden tener como usuarios a estudiantes –en primera instancia– y docentes de lenguas extranjeras, así como servir de base para el desarrollo de aplicaciones para el aprendizaje de lenguas asistido por ordenador (Raine, 2018). En una situación de producción en una lengua extranjera, un estudiante, dado un deficit informativo o inseguridad con respecto a una estructura argumental concreta, puede manejar el recurso según tres finalidades centrales:

i.    informativo-descriptiva: el usuario recaba información preestablecida sobre la estructura concreta que selecciona y consulta;

ii.   tentativa-consultiva: el usuario intenta generar su propia frase y comprueba si está consignada como adecuada en el recurso;

iii.  tentativa-tutorizada: el usuario realiza ejercicios diseñados para la práctica de la valencia.

Se trata, en definitiva, de un modelo para un nuevo tipo de diccionario de combinatoria, que pone el foco en el usuario de los recursos digitales y la individualización de los contenidos que demandan (véase también Gouws, 2017: 26):

> [W]hat is needed is not only a lexicographical tool that is capable of dealing with types of users and situations, but one that provides the necessary mechanisms for individualization of dictionary content – in terms of customizing the views that an individual user is given on the lexicographic data (Spohr, 2011: 103)

## 4. De la interfaz sintáctico-semántica a la generación automática

### 4.1. El esquema argumental como punto de partida

Los prototipos de generación automática ilustran esquemas sintáctico-semántico monoargumentales del nombre –*Xera*– y biargumentales –*Combinatoria*. Dado que la unidad objeto de estudio es el esquema argumental, tenemos que establecerlo para cada sustantivo en cada una de las lenguas, procedimiento que ejemplificamos con el conjunto nominal *dolor | Schmerz | douleur* en su acepción 'Sensación molesta de tipo físico' (Figuras 7, 8 y 9)**.** Las tres lenguas recurren, como era de esperar, a diferentes realizaciones formales, si bien comparten un mismo esquema triargumental formado por un Argumento[1] –el cual desempeña el rol semántico 'Aquel que experimenta el dolor'–, un Argumento[2] –'Aquello que es el origen del dolor'– y un Argumento[3] –'Lugar en donde se localiza el dolor'**:**

Figura 7. Estructura argumental y realizaciones de esp. *dolor*

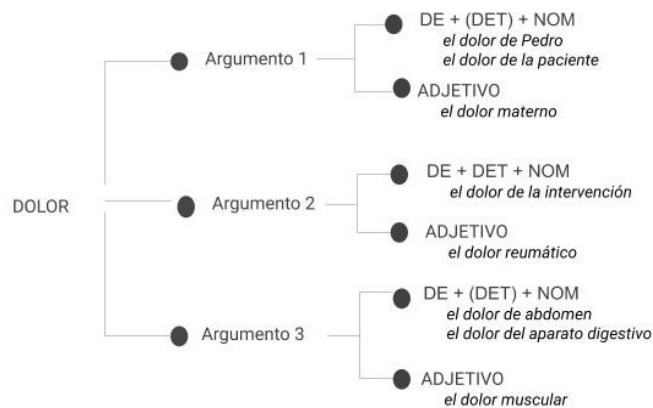

Figura 8. Estructura argumental y realizaciones de fr. *douleur*

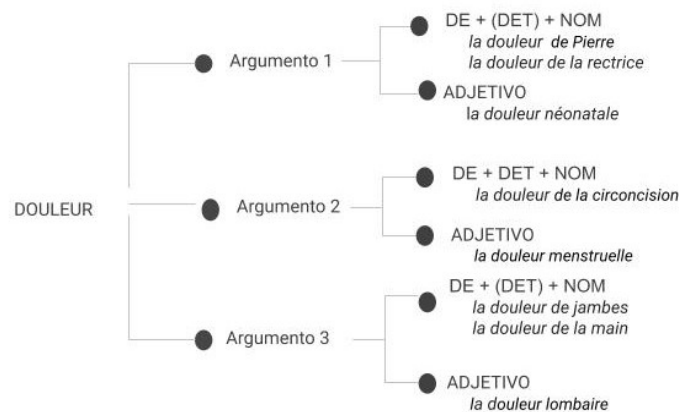

Figura 9. Estructura argumental y realizaciones de dt. *Schmerz*

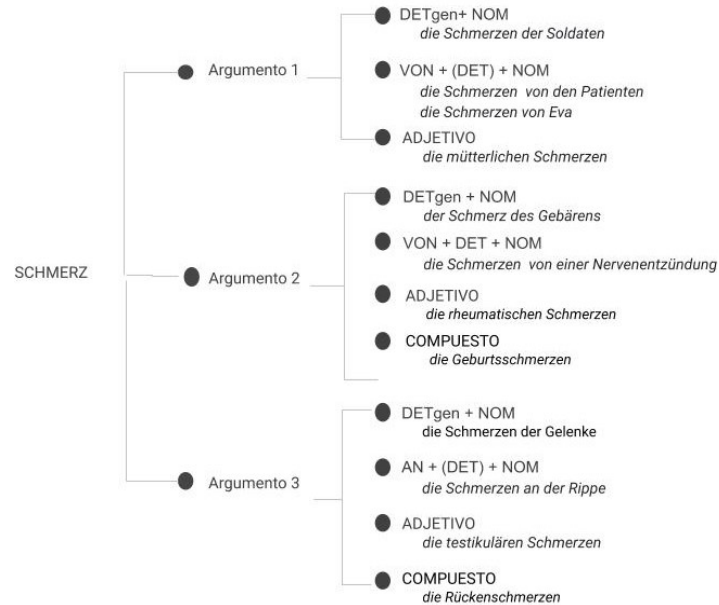

Para garantizar el adecuado funcionamiento de las herramientas, es necesario, además, codificar no solo regularidades –como el hecho de que algunos núcleos solo cuenten con número singular (dt. Anwesenheit, `presencia´, frente a otros con ambos (esp. texto)–, sino también una serie de incompatibilidades y preferencias formales y semánticas de la frase nominal, como, por ejemplo:

## I. Restricciones formal-argumentales

a) Núcleos en singular o plural: la configuración argumental determina algunas restricciones entre el núcleo en singular y/o plural y los respectivos argumentos:

| (4) | *la muerte del niño* | (7) | *la muerte del niño por intoxicación alimentaria* |
|---|---|---|---|
| (5) | *las muertes de los niños* | (8) | *las muertes de los niños por intoxicación alimentaria* |
| (6) | *\*las muertes del niño* | (9) | *las muertes de los niños por intoxicaciones alimentarias* |
| | | (10) | *\*la muerte del niño por intoxicaciones alimentarias* |

En algunas ocasiones, la presencia del núcleo en singular o plural depende de la dotación semántica del lexema que lo acompañe. Así, el sustantivo fr. *odeur* suele aparecer en plural al realizarse en una estructura del tipo ODEUR + DE + {lugar población ciudad}, por ejemplo: *Les odeurs de Paris*.

b) Argumentos en singular o plural: algunos argumentos, como en (11), se realizan en plural, frente a otros en singular (12):

| (11) | *El dolor de huesos* | (12) | *El dolor de cabeza | cuello | espalda | barriga* |
|---|---|---|---|

La selección del número no se explica, en este caso, mediante nuestro conocimiento enciclopédico –solo tenemos un cuello o una barriga–, puesto que la parte del cuerpo también aparece expresada en singular con un experimentante en plural, como, por ejemplo:

(13) *El dolor de cabeza de los enfermos*

## II. Determinantes

a) Su presencia o ausencia va ligada en algunos sustantivos con la expresión de un rol concreto. En (14) la parte del cuerpo en donde se localiza el dolor carece de artículo. Su inclusión reconfiguraría la estructura argumental, de modo que la frase preposicional «de la enferma» (ej. 15) pasaría a ser un modificador de la parte del cuerpo:

| (14) | *el dolor de dientes de la enferma* | (15) | *el dolor del diente molar de la enferma* |
|---|---|---|---|

b) Determinación frente a indeterminación: existen numerosos casos que ejemplifican la necesidad de actualizar un determinante determinado o uno indeterminado:

| | |
|---|---|
| (16) *Die Frage der Arbeitlosigkeit*<br>[La cuestión del paro] | (17) *\*Die Frage einer Arbeitlosigkeit*<br>[*La cuestión de un paro] |

## III. Combinatoria argumental sintagmática

a) La interdependencia entre roles determina la configuración de la estructura argumental. Así, por ejemplo, (18) es agramatical, frente a (19); una casuística semejante se observa con la expresión del rol expansivo en los ejemplos (20)-(22):

| | |
|---|---|
| (18) *\*Die Flucht von Madrid*<br>[La huida de Madrid] | (20) *Der Aufenthalt von 3 Tagen*<br>[La estancia de 3 días] |
| (19) *Die Flucht* von *Madrid nach Santiago*<br>[La huida de Madrid a Santiago] | (21) *Der Aufenthalt von November bis Dezember*<br>[La estancia de noviembre a diciembre] |
| | (22) *\*Der Aufenthalt von November*<br>[*La estancia de noviembre] |

b) La inclusión de adjetivos o adverbios presupone también un estudio de sus restricciones formales y semánticas en el eje sintagmático, y esto, a su vez, en diferentes planos: en (23) parece ser poco común asociar el olor de excrementos a algo agradable, en (24) existe una contradicción debida a la aparición de dos adjetivos que significan lo contrario:

| | |
|---|---|
| (23) *\*El agradable olor a excrementos es intenso* | (24) *\*El agradable olor a excrementos resulta desagradable.* |

Estos ejemplos muestran una vez más que no solo el significado combinatorio (relacional y categorial) es relevante para el estudio de la frase y la oración, sino también la aparición (o no) de un contexto más amplio.

### 4.2. El significado combinatorio

Como recogen las figuras 7-9 y, además es bien sabido, una forma puede expresar diferentes significados, así como a la inversa, un mismo significado puede ser materializado mediante formas diversas. Dado que el objetivo de nuestras herramientas es la generación automática de ejemplos y estructuras nominales no solo gramaticalmente correctos, sino semánticamente adecuados, nuestro modelo presta especial atención a la descripción del significado combinatorio –frente al significado inherente– que comprende, por una parte, el significado relacional (roles semánticos, *vid.* 4.3.) y, por otra, el significado categorial, esto es, las categorías ontológicas que pueden ocupar determinadas casillas funcionales (*vid.* 4.4.).

Para la descripción del significado relacional (Tabla 1) recurrimos a 4 roles semánticos centrales –*Agentivo, Afectado, Clasificativo* y *Locativo*– y su indexación, tal y como propone Engel (2004) y, posteriormente, Domínguez y Valcárcel (2020) para estudios contrastivos:

Tabla 1. Roles semánticos didactizados - Portlex

| Agentivos y Afectados | Clasificativos y Situativos |
|---|---|
| *Aquel/aquello que realiza una acción* | *Clasificativo* |
| *Aquel/aquello afectado* | *Extensión* |
| *Aquel/aquello no afectado* | *Situación: Locación y Tiempo* |
| *Aquel/aquello no afectado: Tema* | *Locación: Origen* |
| *Aquel/aquello que tiene o dispone de algo* | *Locación: Paso* |
| *Aquel/aquello que experimenta un estado* | *Locación: Dirección* |
| *Aquel/aquello que experimenta un nuevo estado* | |
| *Aquel/aquello que existe o es* | |
| *Aquel/aquello que resulta/comienza/finaliza* | |
| *Aquel/aquello que es el origen o la causa* | |
| *Aquel/aquello que es una localización abstracta* | |
| *Aquel/Aquello que es objetivo no espacial* | |

Este inventario de roles didactizados, ya contrastado en el diccionario multilingüe Portlex (2018), permite el análisis lingüístico del material expresivo: a) sustenta la adscripción de los sustantivos a diferentes escenarios (vid. 2.2.), por tanto, no es arbitraria (Engel, 1996: 227). Mediante la aplicación de los roles semánticos diferenciamos, por ejemplo, una frase preposicional que expresa un rol afectado resultativo –*La construcción del edificio*– frente a una misma realización que representa un rol afectado mutativo –*El aumento del porcentaje*; b) los roles contribuyen a diferenciar diferentes tipos de estructuras argumentales, como es el caso de esp. *huida* –que puede expresar el lugar por donde se huye–, frente a esp. *mudanza* –que no lo contempla en su patrón argumental. Por tanto, la descripción de los roles semánticos y su interacción es central para la propia delimitación de la estructura argumental, así como para la desambiguación o agrupación de significados (*vid*. 4.3.).

Dado que una casilla semántico-argumental puede estar ocupada por una o varias categorías ontológicas, la descripción del significado categorial –como parte del significado combinatorio– es otro de los pilares de los prototipos. Nuestra ontología parte de las categorías aplicadas en la gramática y lexicografía valencial, tales como [humano], [objeto], [situación], etc., y evoluciona hacia un aparato descriptivo más amplio y detallado (Domínguez/Valcárcel/Bardanca, 2021). Esta subcategorización de rasgos categoriales generales se requiere ya no solo desde un punto de vista descriptivo-lingüístico, sino también dada la necesidad de codificar la red de combinaciones semánticas categoriales y relacionales más relevantes para su posterior generación (*vid*. 4.4.).

### 4.3. Los roles semánticos

#### 4.3.1. Los roles semánticos como desambiguadores de significado

Un análisis centrado en el plano relacional –en los roles semánticos como instancias organizadoras del material expresivo– contribuye a diferenciar y organizar realizaciones formales en las diferentes lenguas de análisis, aspecto fundamental a la hora de generar frases. Veamos tres casos relativos a la importancia de aplicar los roles semánticos en nuestro modelo.

a)  Adjudicación de ejemplos a diferentes acepciones de significado: los datos de frecuencia y coaparición extraídos de *Sketch Engine* mediante búsquedas CQL (*Corpus Query Language*) no se pueden implementar directamente en nuestros generadores sin un filtrado semántico previo. Este es el que nos permite, por ejemplo, determinar que el primer elemento del compuesto en *Laternenumzug* y *Wohungsumzug* expresa roles semánticos diferentes: en el primer caso un *Clasificativo*, en el segundo un *Locativo*. Dicho análisis posibilita, por tanto, asignar el primer ejemplo a una acepción de significado del tipo 'desfile', frente al segundo caso, 'mudanza'.

b)  Atribución de representantes de una misma clase de palabra a diferentes funciones sintáctico-semánticas: dado que mediante la búsqueda en corpus no se pueden delimitar adjetivos en función de complementos específicos (esp. *dolor estomacal*) frente a los no específicos (esp. *fuerte dolor* o *dolor crónico*), ni tampoco diferenciar tipos de argumentos específicos, tras la evaluación cuantitativa (Tab. 2 a modo de ejemplo; datos de *Sketch Engine*) se precisa un análisis semántico-relacional.

Tabla 2. Adjetivos pre y postnominales más frecuentes con el sustantivo esp. *dolor*.

|  | prenominales | postnominales |
|---|---|---|
| **Coapariciones** | 94 718 | 225 423 |
| Prototipos | fuerte (17 801) grande (12 233) | crónico (20 514) físico (10 450) |

A partir del análisis de los roles semánticos, se concluye que i) los 100 primeros adjetivos prenominales son modificadores no específicos, por tanto, no representan realizaciones de roles semánticos – por tanto, *fuerte, grande, crónico* y *físico* no pueden ser incluidos como ejemplos de realizaciones valenciales de tipo adjetival – y ii) entre los 100 primeros adjetivos postnominales un 47 % son no específicos, un 45 % expresan el Argumento[3] (Locación), el 7 % el Argumento[2] (*Origen*) y solo un 1 % representa el Argumento[1] (*Experimentante*). La figura 10 presenta lo anteriormente expuesto:

Figura 10. Análisis semántico de esp. (adjetivo) + *dolor* + (adjetivo)

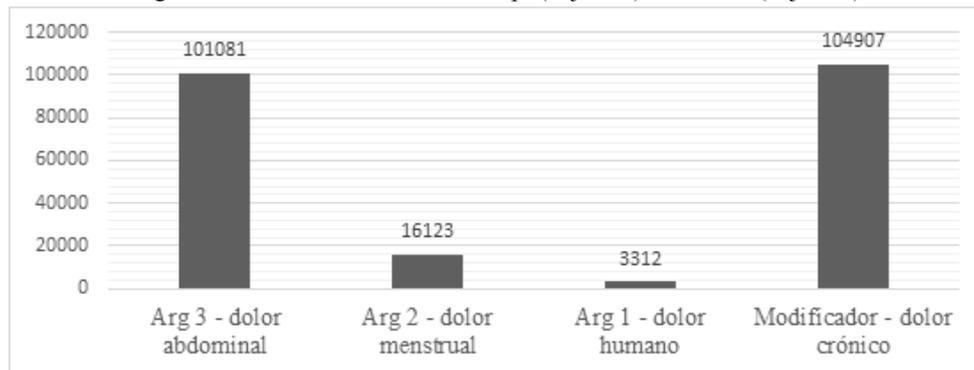

c)   Clasificación de una estructura formal atendiendo a diferentes roles: es comúnmente conocido que las lenguas románicas cuentan con preposiciones polifuncionales, como, por ejemplo, la preposición esp. *de* o fr. *de*, al igual que sucede con el abanico de relaciones que puede expresar el caso genitivo en alemán (véanse las figuras 7-9). Nuevamente, el análisis de los roles semánticos y su combinatoria sirve de filtro para determinar la combinatoria sintáctico-semántica de los diferentes argumentos implicados en la estructura valencial, por tanto, la relación entre formas, significados y clases semánticas prototípicas, tal y como recoge la Tabla 3:

Tabla 3. Descripción semántica de la metaestructura **[DE+ DET+NOM]** del sustantivo esp. *dolor*.

| Argumento 1 | DE + DET + NOM | |
|---|---|---|
| Rol semántico:<br>Experimentante | 1. | animado animal mamífero - *el {intenso} dolor {intenso} del animal de carga* |
| | 2. | animado humano cargo - *el {repentino} dolor {repentino} de la diplomática* |
| | 3. | animado humano profesión - *el {agudo} dolor {agudo} del soldador* |
| | 4. | animado humano condición humana - *el {leve} dolor {leve} de los competidores* |
| | 5. | animado humano condición humana negativa - *los {insoportables} dolores {insoportables} de la enferma* |
| **Argumento 2** | **DE + DET + NOM** | |
| Rol semántico:<br>Origen o causa | 1. | proceso humano médico - *el {agudo} dolor {agudo} de la intervención* |
| | 2. | proceso humano violencia - *el {intenso} dolor {intenso} de la pelea* |
| | 3. | proceso natural biológico - *el {lacerante} dolor {lacerante} del alumbramiento* |
| | 4. | proceso cambio motricidad - *el {intenso} dolor {intenso} del golpe* |
| | 5. | proceso natural patológico - *el {insoportable} dolor {insoportable} de la peritonitis* |
| **Argumento 3** | **DE + DET + NOM** | |
| Rol semántico: Locación | 1. | animado humano órgano - *el {agudo} dolor {agudo} del aparato digestivo* |
| | 2. | animado animal parte del cuerpo - *el {repentino} dolor {repentino} del maxilar* |
| | 3. | animado humano parte del cuerpo - *el {ligero} dolor {ligero} del antebrazo* |

La detección del rol semántico *Locación* en dicha metaestructura permite asignar dichas realizaciones a la acepción 'sensación molesta de tipo físico' (y no a otra acepción de *dolor*, la relacionada con el sentimiento).

Esta aproximación desde la semántica puede ser una futura vía de análisis automático en favor de la desambiguación de datos extraídos de corpus. A su vez, sin un análisis semántico relacional no es posible describir y, posteriormente, codificar las realizaciones adjetivales que expresan roles semánticos y, por ende, argumentos valenciales.

### 4.3.2. Los roles semánticos como instancias grupales

Dado que el criterio aplicado para establecer el catálogo de complementos nominales atiende a que los mismos sean específicos de un portador valencial, resulta evidente que el tipo de realización formal no puede influir en el análisis. Por tanto, en nuestros generadores se incluyen aquellas formas que muestran una relación paradigmática entre sí desempeñando el mismo rol. Así, los esquemas argumentales contemplan no solo frases en genitivo y frases con *von* (para el alemán), frases preposicionales, sino también adjetivos, compuestos y estructuras apositivas [N1 + N2] (Domínguez y Valcárcel, 2020; Valcárcel, 2017). Es nuevamente el significado relacional el común denominador para el análisis del material expresivo. Veamos algunos ejemplos:

El sustantivo dt. *Frage ('pregunta')* puede expresar mediante el genitivo y realizaciones preposicionales el rol Agente –'Aquel que plantea la pregunta' (25)– y mediante una frase preposicional con *nach* 'Aquello sobre lo que se plantea la pregunta' (27). Nada impide entender, atendiendo a la especificidad y la relación paradigmática señalada anteriormente, la primera parte de los compuestos de *-frage* como posibles realizaciones de agente (26) o de afectado (28):

| | | | |
|---|---|---|---|
| (25) | *die {lustige} Frage der Studentin* <br> [la {divertida} pregunta de la estudiante] | (26) | *die {interessante} Teilnehmerfrage* <br> [la {interesante} pregunta del participante] |
| (27) | *die {unerwartete} Frage nach dem Ergebnis* <br> [la {inesperada} pregunta sobre el resultado] | (28) | *die {wichtige} Zukunftsfrage* <br> [la {importante} pregunta sobre el futuro] |

Este mismo procedimiento subyace al análisis de las expresiones apositivas [N1 + N2], como en *Geschmack Schokolade | sabor chocolate | saveur chocolat* (Valcárcel, 2017:202). Su carácter argumental se sustenta nuevamente en su especificidad: así, mientras sustantivos como dt. *Flucht* no contemplan una [N1 + N2] en su esquema argumental, en otros, como dt. *Aufenthalt*, sí que es el caso, como muestran las siguientes realizaciones formales del rol semántico expansivo o dilativo:

(29) *der Aufenthalt von 2 Tagen*
   [la estancia de 2 días ]
(30) *der zweitätige Aufenthalt*
   [la de dos días estancia]
(31) *der Jahresaufenthalt*
   [la de 1 año+estancia]
(32) *2 Tage Aufenthalt*
   [2 días estancia]

### 4.3.3. A modo de resumen

Los roles semánticos actúan de instancias organizadoras del material expresivo, ya no solo desde un punto de vista monolingüe sino también multilingüe, pudiendo operar, por tanto, de *tertium comparationis*. Junto con su papel central a la hora de desambiguar significados, es el análisis relacional –como parte del significado combinatorio– un filtro imprescindible para determinar la interfaz sintáctico-semántica. Tanto los roles semánticos como el significado categorial nos permiten establecer prototipos léxicos y clases semánticas prototípicas (*vid.* 4.4.), fundamentales para la generación automática.

### 4.4. Rasgos ontológicos como componentes del significado combinatorio: prototipos léxicos y clases semánticas

Tras el análisis cuantitativo y semántico-relacional, se continúa filtrando y categorizando el caudal léxico atendiendo a su significado categorial. Mediante este procedimiento se consigue recabar información sobre qué candidatos léxicos – entidades ontológicas– pueden aparecer en qué casillas argumentales, lo cual es fundamental para describir y codificar la combinatoria argumental para su posterior generación automática. Para tal fin, recurrimos a un concepto propio de prototipo léxico –candidato representativo de una casilla valencial (*vid.* Domínguez, 2021)–, como es el caso de {cabeza} o {espalda} en la Tabla 4. A partir de este prototipado léxico, esto es, de esta categorización ontológica, estamos en disposición de establecer las así denominadas clases semánticas prototípicas (Domínguez, Solla y Valcárcel, 2019):

Tabla 4. Prototipos y prototipado léxico en la estructura [DOLOR + DE + NOM: Arg3: Locación del dolor]

| prototipos | posición | coapariciones | | prototipado léxico |
|---|---|---|---|---|
| cabeza | 1 | 147 678 | | parte del cuerpo externa |
| espalda | 2 | 29 719 | | parte del cuerpo externa |
| cuello | 7 | 3840 | animado humano | parte del cuerpo externa |
| ovario | 13 | 1491 | | órgano |
| hueso | 14 | 1.484 | | parte del cuerpo interna músculos y huesos |

Por tanto, siguiendo este procedimiento agrupamos los prototipos léxicos de una estructura argumental en clases semánticas, o lo que Gross (2008, 11) denomina una «classe d´objets»:

> "un ensemble de substantifs, sémantiquement homogènes, qui détermine une rupture d'interprétation d'un prédicat donné, en délimitant un emploi spécifique. Cette définition implique que les classes d'objets ne sont pas des concepts sémantiques abstraits mais des entités construites sur des bases syntaxiques et déterminées par la signification des prédicats"

Ambos, prototipos léxicos y clases semánticas, refieren a una estructura argumental de un portador valencial concreto, así como a la concreta actualización de su significado en una acepción determinada. Esto es, los prototipos y los candidatos léxicos recogidos en cada clase semántica no son categorías universales, sino que su categorización es unívoca y ligada a un sustantivo concreto. Sirve como ejemplo la Tabla 5.

Tabla 5. Candidatos léxicos y clase semántica desde una perspectiva comparativa.

| Patrón argumental | Clase semántica | Candidatos léxicos |
|---|---|---|
| [determinante + MUERTE + DE + Argumento] | {estado físico} | \| *desnutrición* \| *enfermedad* \| *hambre* \| *inanición* \| *sed* \| *la muerte de salud*, *la muerte de líbido* |
| [determinante + TEXTO + SOBRE + determinante + Argumento] | {estado físico} | \| *desnutrición* \| *enfermedad* \| *hambre* \| *inanición* \| *sed* \| *líbido* \| *salud* |

La ontología que da sustento a los generadores (Domínguez, Valcárcel y Bardanca, 2021) contempla la clase semántica {estado físico}, que se observa en determinadas estructuras de los sustantivos *muerte* y *texto*. La lista de candidatos léxicos que conforman dicha clase semántica para cada uno de estos sustantivos no es coincidente en su totalidad: tal y como se observa en la Tabla 5, algunos de los representantes léxicos en el sustantivo *texto* no lo son en el caso de *muerte*: *la muerte de salud*, *la muerte de líbido*. Por tanto, una clase semántica cuenta con diferentes representantes léxicos según el sustantivo objeto de descripción o consulta.

Cabe señalar aún tres cuestiones importantes:

a) Los prototipos léxicos y las clases semánticas sientan las bases para la asignación del léxico a las acepciones de significado y, por tanto, para la desambiguación de significados (*vid*. 4.3.1.). Así, por ejemplo, una estructura argumental del tipo DOLOR + DE + NOM [Origen] contempla entre sus clases semánticas diferentes tipos de procesos con sus respectivos candidatos:
{proceso natural biológico}: *el dolor del alumbramiento* \| *parto* \| *ovulación*
{proceso natural patológico}: *el dolor de la peritonitis* \| *neuralgia* \| *cólico*
Cualquier otro candidato léxico intercambiable con los anteriores en el eje paradigmático –compartiendo la misma clase y rol semántico– ha de pertenecer, por tanto, a la misma acepción de significado que los anteriores.

b) Otra de las aportaciones de la aplicación de las clases semánticas tiene que ver con el usuario, puesto que le permite, por ejemplo, observar la actualización preposicional y su alternancia:

Tabla 6. Clases semánticas en la estructura determinante + PREGUNTA + preposición+ NOMBRE [Tema]

| pregunta +de + Nombre | pregunta + sobre + Nombre |
|---|---|
| 1.  intelectual área de conocimiento<br>las {interesantes} preguntas {interesantes} de anatomía | 1.  intelectual área de conocimiento<br>las {incomprensibles} preguntas {incomprensibles} sobre aerodinámica<br>2.  intelectual contenido general<br>las {inesperadas} preguntas {inesperadas} sobre reglas gramaticales<br>3.  lugar población ciudad nombre propio<br>las {innumerables} preguntas {innumerables} sobre Montevideo<br>4.  animado criatura de ficción nombre propio<br>las {impredecibles} preguntas {impredecibles} sobre Don Quijote<br>5.  lugar población país nombre propio<br>las {interesantes} preguntas {interesantes} sobre Brasil<br>6.  animado humano personaje histórico<br>la {interesante} pregunta {interesante} sobre Marco Polo |

Como se observa en la Tabla 6, en la herramienta *Xera* las clases semánticas vienen acompañadas de ejemplos estándar, los cuales ilustran una de las posibles realizaciones de la clase semántica a la que refieren. En los mismos se presentan entre corchetes adjetivos modificadores en posición pre y postnominal para el esp. y fr, y en posición prenominal para el alemán. Estos resultan del análisis de listas de frecuencia extraídas de *Sketch Engine*.

c) Junto con la contribución del significado categorial al análisis lingüístico-semántico de la frase nominal, es importante recordar que la expansión léxica en el eje paradigmático se articula siguiendo criterios ontológicos (*vid*. 2.2.). Dicha expansión proporciona a los generadores variación formal de candidatos léxicos para la expresión de diferentes roles y clases semánticas. Es este aspecto, junto con la coherencia semántica y la fluidez, un factor especialmente relevante en la evaluación de los generadores de la lengua natural (*vid*. Hashimoto, Zhang y Liang, 2019).

## 5. Conclusión y perspectivas de futuro

Los generadores cumplen con el propósito principal para el que han sido concebidos: servir de prototipos lexicográficos de generación plurilingüe automática de patrones argumentales del nombre y sus ejemplos.

La propuesta metodológica diseñada para los simuladores, que conjuga diferentes aproximaciones lingüístico-lexicográficas, recursos –como WordNet–, así como técnicas y procedimientos de las áreas del PLN y GLN se ha verificado, de tal modo que *Xera*, *Combinatoria* y *CombiContext* están actualmente en funcionamiento. Además, esta metodología conjugada ya ha sido extrapolada para el diseño de nuevos recursos, como es el caso de *XeraWord* (2021), un generador probeta para el gallego y portugués. Esta herramienta, presentada en Domínguez y Caíña (2021), muestra similitudes metodológicas con la herramienta *Xera*, pero también notables diferencias: (a) *XeraWord* permite una aproximación ontológico-conceptual a los patrones monoargumentales; (b) Para la obtención de datos léxicos este generador recurre además a la traducción automática. Una explicación más detallada sobre dicha herramienta y su aplicación para estudios comparativos entre el gallego y portugués, así como entre estos y otras lenguas, se encuentra en el trabajo citado previamente.

Los generadores han dado lugar también al diseño de herramientas *ad hoc* para el análisis lingüístico (Domínguez, Solla y Valcárcel, 2019), así como, en el marco de *XeraWord*, a un traductor – *TraduWord*– del caudal léxico extraído de WordNet. Las futuras vías de análisis y estudio científico no se agotan aquí, puesto que la aproximación semántica adoptada en la concepción de los generadores puede ser un punto de partida para la desambiguación automatizada de datos de corpus.

Demostrada la viabilidad metodológica de los prototipos, las herramientas se están optimizando en diferentes niveles, para con ello facilitar su aplicación como recursos lexicográficos en la enseñanza-aprendizaje digital de lenguas y/o nutrir a otros recursos con ejemplos de combinatoria:

a) Desde un punto de vista cuantitativo: Continuamos evaluando diferentes procesos de automatización en los procedimientos, de modo que el número de unidades de análisis se pueda aumentar significativamente en un tiempo menor. La inclusión de nuevas lenguas – como el caso de *XeraWord* –es otra vía futura de trabajo.

b) Desde un punto de vista cualitativo:
   i. Actualmente los recursos se están optimizando con datos relativos de datos relativos al contexto frasal y al marco oracional.
   ii. Es necesario diseñar diferentes modos de acceso a la información y ampliar las finalidades de contexto interno de uso, así como la didactización de ciertas etiquetas. Esta focalización del usuario del recurso es tal que actualmente los prototipos se están rediseñando para ofrecer también un acceso onomasiológico a los datos, tal y como ya se ha aplicado en el recurso *XeraWord*.

Los simuladores son una firme apuesta para un nuevo modelo de diccionario de valencias, que permite la consulta y la interacción, de modo que hacemos nuestra la célebre expresión de Klein (2015, 294) «das Wörterbuch der Zukunft ist kein Wörterbuch, es ist mehr» (`El diccionario del futuro no es un diccionario, es algo más.´). Pero ya no solo, puesto que el volumen de datos etiquetados sobre el comportamiento de la frase nominal no es nada desdeñable, como tampoco lo es la posible aplicación de estos recursos como léxicos computacionales e integrables en otras herramientas. En definitiva, un recorrido por este trabajo nos permite observar un modelo para desarrollar nuevos instrumentos lexicográficos a partir de conocimientos lexicográficos existentes mediante una apuesta decidida por la retroalimentación de recursos y por una mayor interacción entre la lexicografía y áreas del procesamiento y generación del lenguaje natural.

## Agradecimientos

## Referencias

*xeración automática da argumentación da frase nominal en galego e portugués*. Santiago de Compostela: Instituto da Lingua Galega. http://ilg.usc.gal/xeraword/

## Otros recursos electrónicos y diccionarios

AnCora = http://clic.ub.edu/corpus/es/ancora
AncoraNom = http://clic.ub.edu/corpus/es/ancoranom_es
CPA = http://www.pdev.org.uk/
Framenet = https://framenet.icsi.berkeley.edu/fndrupal/
Kind = http://www.tecling.com/cgi-bin/kind/2020/
NomBank = https://nlp.cs.nyu.edu/meyers/NomBank.html
OBELEXmeta = https://www.owid.de/obelex/meta?info
Portal lexicográfico PORTLEX = http://portlex.usc.gal/
Propbank = http://verbs.colorado.edu/propbank/framesets-english-aliases/
Sketch Engine = https://www.sketchengine.eu
Verbario = http://www.verbario.com
Verbnet = https://verbs.colorado.edu/~mpalmer/projects/verbnet.html